\documentclass[runningheads]{llncs}

\usepackage[utf8]{inputenc}
\usepackage[english,vietnamese]{babel}
\usepackage{graphicx}
\usepackage{indentfirst}
\usepackage{ulem}
\usepackage{multirow}
\usepackage[T1]{fontenc}
\usepackage{makecell}
\usepackage{fixltx2e}
\usepackage{amsmath}
\newcolumntype{P}[1]{>{\centering\arraybackslash}p{#1}}
\begin{document}
\selectlanguage{english}
\title{Hierarchical Transformer Encoders for Vietnamese Spelling Correction}
%
%
\author{
Hieu Tran\textsuperscript{1,2} \and
Cuong V. Dinh\textsuperscript{1} \and
Long Phan\textsuperscript{1} \and
Son T. Nguyen\textsuperscript{1,2,3}}
\authorrunning{Tran et al.}
%
\institute{\textsuperscript{1}Zalo Group - VNG Corporation, Ho Chi Minh City, Vietnam\\
\textsuperscript{2}University of Science, Ho Chi Minh city, Vietnam.\\
\textsuperscript{3} Vietnam National University, Ho Chi Minh city, Vietnam.\\
\email{hieutt9@vng.com.vn, ntson@fit.hcmus.edu.vn}}
\maketitle              
\begin{abstract}
In this paper, we propose a Hierarchical Transformer model for Vietnamese spelling correction problem. The model consists of multiple Transformer encoders and utilizes both character-level and word-level to detect errors and make corrections. In addition, to facilitate future work in Vietnamese spelling correction tasks, we propose a realistic dataset collected from real-life texts for the problem. We compare our method with other methods and publicly available systems. The proposed method outperforms all of the contemporary methods in terms of recall, precision, and f1-score. A demo version\footnote{https://nlp.laban.vn/wiki/spelling\_checker/} 
is publicly available.
\end{abstract}

\section{Introduction}
Spelling correction has always been a practical problem with many real-world applications. Popular word processor applications such as Microsoft Word, Google Docs, LibreOffice Writer, specialized applications such as Grammarly, or other text-support platforms often integrate spelling correction systems to improve their users' productivity in typing and editing documents. Meanwhile, most spoken languages such as English or Chinese are thoroughly researched in the problem, there are only a few studies on Vietnamese language. This leads to a very poor user experience for the people. Therefore, as the aim of this paper, we would like to develop a method to detect and correct spelling mistakes for Vietnamese. This method is integrated into our products to help our news authors and messenger users to check their writings.

With the improvement in computational power, recent studies \cite{Zhang2020SpellingEC,Niranjan2020HierarchicalAT,korean} prefer machine learning methods over rule-based methods in the spelling correction problem, especially the introduction of Transformer architecture \cite{Vaswani2017AttentionIA} and BERT \cite{Devlin2019BERTPO} have boosted not only this problem but a lot of other problems in natural language processing. Not out of the mainstream, specifically for Vietnamese, we design a novel hierarchical neural network model that inspired by contemporary methods in the field. Our model consists of multiple Transformer encoders that encode the input text in both character-level and word-level simultaneously to detect errors and make corrections.

Furthermore, to the best of our knowledge, currently, there are no common, realistic test sets for Vietnamese that research community can compare their works. Studies in Vietnamese \cite{DBLP:journals/corr/abs-1709-07104} (and some other languages) often artificially generate their test sets based on their own analysis of common mistakes. Although this approach is inexpensive and convenient in research environment, we highly doubt its practicality while using industrial products. Therefore, we propose a test set that highly reflects reality. The data are manually collected by examining Vietnamese Wikipedia drafts, which are usually submitted without much spelling revision and so represent the behaviors in the Vietnamese spelling problem.

In total, this paper offers the following contributions:
\begin{itemize}
\item A novel method that applies a hierarchical Transformer model with character-level and word-level encoders for Vietnamese spelling detection and correction;
\item We propose a Vietnamese spelling correction test set for comparison with future studies; 
\item Experimental results with various methods show that the quality of Vietnamese detecting and error correction is promising and usable in industry.
\end{itemize}

\section{Related Work}

There have been many studies on the spelling correction problem. In \cite{Zhang2020SpellingEC}, the authors propose a deep model called Soft-Masked BERT for Chinese Spelling error correction. The model is composed of two neural networks, detector, and corrector. The detector network based on bidirectional GRUs is to predict the probability of each token being an error. The correction network is fine-tuned from a pre-trained bidirectional Transformer - BERT \cite{Devlin2019BERTPO} to predict the best alternative for the detected token. In the network, the error embeddings are softly masked by summing its original word embeddings and the [MASK] token embedding with the detector's output probabilities as weights. The two networks are jointly trained to minimize their objectives. The model is proved to outperform other architectures in our setting.

In \cite{Niranjan2020HierarchicalAT}, the authors propose a variation of the conventional Transformer that composed of three encoders and a decoder for syntactic spell correction. The three encoders first extract contextual features at character level using unigram, bigram, and trigram, correspondingly, then the decoder processes information from encoders to generate the correct text. All of the encoders have the same architecture as \cite{Vaswani2017AttentionIA}, but the decoder is modified a bit by replacing the one encoder-decoder multi-head attention layer in \cite{Vaswani2017AttentionIA} with three encoder-decoder ones, each one takes outputs from an encoder as input. The entire network is trained to generate the sequence of corrected tokens. 

Similarly, \cite{korean} investigates Korean spelling error correction as a machine translation problem in which an error text is translated to its corrected form. The authors employ common neural machine translation models such as LSTM-Attention, Convolution to Convolution, and the Transformer model. All the models capture context by reading the entire error text through an encoder and translate it by a decoder. 

Recent studies that focus on Vietnamese spelling issue tend to focus on diacritic restoration. There has been an attempt in utilizing phrase-based machine translation and recurrent neural networks in diacritic restoration \cite{DBLP:journals/corr/abs-1709-07104}. The phrase-based machine method tries to maximize the probability from source sentence to target sentence by segmented input into a number of phrases (consecutive words) before being analyzed and decoded; the later method employing a neural network model with the encoder-decoder design, trying to maximize the conditional probability of the target/output sentence. The authors claimed both systems had achieved state-of-the-art for the respective tasks on a data set, in which neural network approach performs at 96.15\% accuracy compared to 97.32\% of the phrase-based one. 

In \cite{4586339}, this work deals with the Vietnamese spelling problem by the detecting phase and the correcting phase. The authors combine a syllable Bi-gram and a Part-of-speech (POS) Bi-gram with some characteristic Vietnamese rules for finding errors. In the correcting phase, Minimum Edit Distance and SoundEx algorithms are applied as a weight function to evaluate heuristic generated suggestions. The significant strength of using the weight function is avoiding evaluating suggestions by specific criteria that can not be suitable for all types of documents and errors.

In earliest work \cite{10.1007/978-981-15-6168-9_40}, instead of focusing on mistyped errors, this study directly cope with the consonant misspelled errors using deep learning approach. The bidirectional stacked LSTM is used to capture the context of the whole sentence, error positions are identified and corrected based on that.

\section{Data}
In this section, we present our method in generating training data and the process of collecting test set for the Vietnamese spelling correction problem.

\subsection{Training data}

The cost of building a big and high-quality training set would be prohibitive. Fortunately, training data for the spelling correction problem can easily be generated through mistake-free texts which are available online and simple analysis of common spelling mistakes. For a mistake-free corpus, we collect news texts published from 2014 to date from various Vietnamese news sources. Those texts were thoroughly revised and spelling checked by the publishers, so they only have a reasonable amount of spelling mistakes. We also include Vietnamese Wikipedia\footnote{https://vi.wikipedia.org/wiki/Wikipedia} articles in the data. In addition, we consider oral texts by collecting high-quality movie subtitles, which only contains spoken language. In total, there are up to 2 GB of news/Wikipedia text data and 1 GB of oral text data.

Vietnamese spelling mistakes can be easily encountered in everyday life. We classify the most common mistakes into three categories: 
\begin{otherlanguage}{vietnamese}
\begin{itemize}
    \item Typos: this type of mistakes is generated while people are typing texts. It includes letter insertions, letter omissions, letter substitutions, transportation, compounding errors, or diacritic errors. Most people type Vietnamese in two styles: VNI and Telex, which use either numbers or distinctive characters to type tones or special characters. For example, "hà nội" (English: Hanoi, the capital city) is often mistyped as "haf nooij" when one forgot to turn on Vietnamese mode or was typing too fast.
    \item Spelling errors: this type of mistakes is due to official conventions or regional dialects in Vietnam. For example, people who live in the north of Vietnam typically have trouble differentiating d/r/gi, tr/ch, s/x, n/l sounds. They often mispronounce or misspell the word "hà nội" to "hà lội" (English: no meaning). This type includes both letter or diacritic errors.
    \item Non-diacritic: although this sometimes is intentional we take the diacritic restoration problem into account. Non-diacritic Vietnamese texts are much faster to type and fairly understandable under specific circumstances and contexts. However, these texts sometimes cause misunderstanding and are very informal. Therefore we would like to offer diacritic restoration to improve further the functionality of the model.
\end{itemize}
During the training data generation, we try to employ a random rule-based generator that covers all above types of spelling errors in diversity over the corpus. At the end, a generated sentence can have a lot of spelling errors or no errors at all. This ensures the model not only learn the core orthography of Vietnamese but also capture the surrounding context to make accurate spelling detection and correction. 
\end{otherlanguage}

\subsection{Wiki spelling test set}

For the most realistic performance measurement of the Vietnamese spelling correction problem, we would like to build up a test set that characterizes general behaviors of spelling mistakes while people are typing. We find out that drafts of Vietnamese Wikipedia articles are often submitted without much spelling revision. Therefore, we collect Wikipedia drafts from varied domains. We select ones that have a substantial number of spelling errors and then manually detect the errors and suggest reasonable substitute words. In total, we revised more than 100 articles with about 1,500 spelling errors and 14,000 sentences (approximately 1,300 sentences with errors). A typical sample in the data set can be found in Table \ref{table:1}.

The dataset is available at GitHub repository\footnote{https://github.com/heraclex12/Viwiki-spelling} under the Attribution 4.0 International (CC BY 4.0) license. The dataset is stored in JSON lines and each document contains the ids, text contents, current revision ids, previous revision ids, Wikipedia page ids, and mistakes.
\begin{table}
\begin{center}
\caption{Some short paragraphs in Wikipedia Spelling Test set}
\begin{otherlanguage}{vietnamese}
\setlength\tabcolsep{1.5pt}
\begin{tabular}{|p{1cm}| @{\hskip5pt}p{5.2cm}@{\hskip5pt} |@{\hskip5pt} p{5.2cm }@{\hskip5pt}|}
    \hline
\thead{No.} & \thead{Content} & \thead{} \\ 
                    \hline
         \centering 1 &  ...TA-50 không đơn thuần là một \sout{\textbf{chuếc}}/\textbf{chiếc} máy bay huấn luyện mà nó còn được đánh giá như một chiến đấu cơ đích thực. Loại chiến đấu cơ này đạt tốc độ \sout{\textbf{tới}}/\textbf{tối} đa lên đến gấp rưỡi vận tốc âm thanh... & ...TA-50 is not only a trainer aircraft but also a fighter aircraft. This type of fighter can reach a maximum speed at 1.5 times the speed of sound...    \\
         & &\\
         \centering 2 & ...Hà Nội là thành phố trực thuộc trung \sout{\textbf{uowng}}/\textbf{ương} có diện tích lớn nhất cả nước từ khi tỉnh Hà Tây \sout{\textbf{xáp}}/\textbf{sáp} nhập vào, đồng thời cũng là địa \sout{\textbf{phuong}}/\textbf{phương} đứng thứ nhì về dân số với hơn 8 triệu người, tuy nhiên... & ..Hanoi has been the largest central city by area in the country since the merging with Ha Tay province; it is also a second most populated city with more than 8 millions people, yet...  \\
         \hline
    \end{tabular}
\end{otherlanguage}
\label{table:1}
\end{center}
\end{table}

\section{Neural Network Architecture}

\begin{figure}
    \centering
    \includegraphics[width=\textwidth]{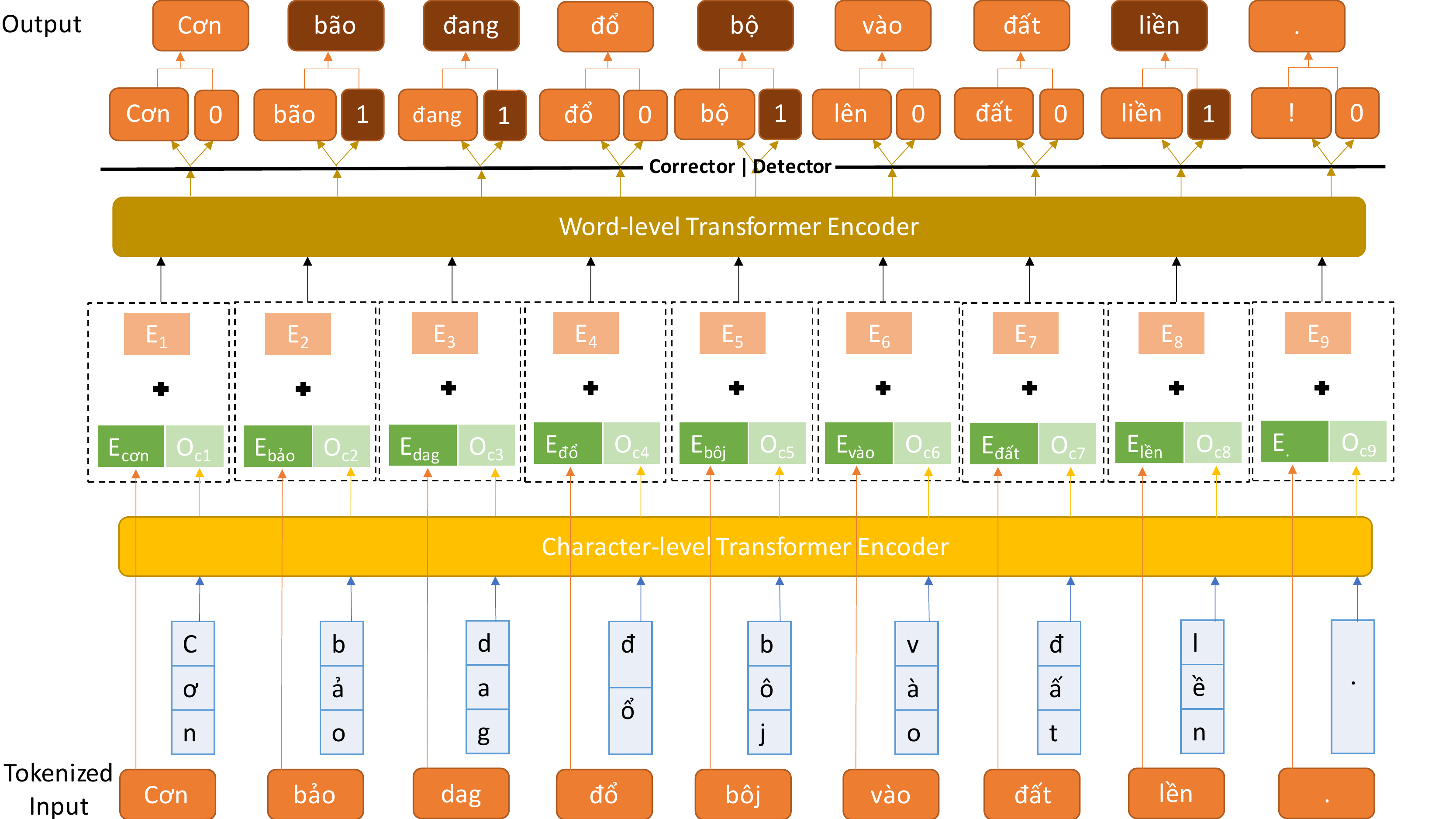}
    \caption{Proposed Neural Network Architecture}
    \label{fig:model}

\small
(Example: The storm is hitting the mainland)
\end{figure}

In this section, we will describe our proposed neural network architectures deliberately designed to detect and correct Vietnamese spelling errors. The network is inspired by the encoder component of the original Transformer \cite{Vaswani2017AttentionIA} in a hierarchical fashion. The model is depicted in Figure \ref{fig:model}.

\subsection{Tokenization}

First of all, the inputted piece of text is broken down into tokens, or says, words based on white-space characters. These tokens are further broken down into characters, and then both these tokens and their characters will be passed on to our model as inputs. Normally, WordPiece \cite{wu2016googles} is a popular Byte Pair Encoding (BPE) sub-word tokenization \cite{10.5555/177910.177914}, which would be used to reduce the exceedingly large amount of vocabulary. However, we do not go for sub-word options for the following reasons. Firstly, Vietnamese is a monosyllabic language whose tokens consist of only one syllable, and that sub-words only stand for phonetic features rather than the meaning. Secondly, Vietnamese has a fairly small amount of tokens. In fact, there are only 7184 unique syllables actually used, and they cover more than 94\% of the content\footnote{http://www.hieuthi.com/blog/2017/04/03/vietnamese-syllables-usage.html}. Lastly, in the spelling error problem, there are a lot of unexpected tokens that may appear due to typos or foreign languages. If sub-word tokenization is applied, resulting out-of-vocab sub-tokens may lead to unknown tokens which either causes loss of character information or redundant computation to deal with too many sub-words. Therefore, we decide to use whitespace tokenization with an additional character-level layer to keep character information. Moreover, we use a special token called UNK to mask out-of-vocab words that are likely to be incorrectly spelled. 

\subsection{Transformer}

For the neural network architecture, at both character and word levels, we adopt Transformer encoder architectures to encode information. The character-level encoder is composed of 4 self-attention layers of hidden size 256 while the word-level one is bigger, with 12 self-attention layers of hidden size 786, which is the same size as conventional BASE-size BERT. Tokenized inputs to the components are vectorized through two embedding tables (i.e. word embedding as $E_{string}$ and character embedding tables as $O$) and are summed with learned positional embeddings $E_{n}$ where $n \in [1, 2,...192]$. In addition, the word embeddings are also concatenated with the corresponding character-level output vectors.

The context and character-aware representations of each word generated by the word-level encoder are used to classify if the token is a mistake and also to predict a substitute in case the token is an actual mistake. The outputting representations are then forwarded to two classifiers, one for detection and the other for correction. Each of two classifiers is composed of two fully-connected layers with the softmax activation function on the last layer. Besides, the correction classifier shares the same weights as the word embedding of the embedding layer. 

After going through the detection classifier, each token of the sequence of tokens is predicted to be incorrect or not. If a token is incorrect, the correction classifier will suggest a suitable replacement.

The objective function for the learning process is the sum of two cross-entropy losses of the two classifiers. As accurate tokens do not need to be corrected, we remove those tokens out of the correction loss function. More clearly, given a sequence of words $x$ as input, the loss function formulated as follows, where $V$ is the vocabulary size; $N$ is the length of sequence of words; $E$ $(E<=N)$ is the number of true spelling mistakes; $y_{ij}$, $p_{ij}$ denote the ground-truth labels and the probability of predictions, respectively.
$$\mathcal{L}=\mathcal{L}_{detector} + \mathcal{L}_{corrector}$$
$$\mathcal{L}=-\frac{1}{N +1e^{-5}}\sum_{i=0}^{N-1}\sum_{j\in \{0; 1\}} y_{ij} \log p_{ij} -\frac{1}{E+1e^{-5}}\sum_{i=0}^{N-1}k_{i}\sum_{j=0}^{V-1} y_{ij} \log p_{ij}$$
$$\text{where} \quad k_i = \begin{cases}1 & \quad \text{if $i$-th token is a spelling mistake} \\ 0 & \quad \text{otherwise}\end{cases}$$

\section{Experiments}
In this section, we briefly introduce our metrics for the detection module and the correction module in the Vietnamese spelling correction problem and then present the experiment settings along with results.

In addition to the Wiki spelling test set, we also evaluate our proposed models on informal corpus which are obtained from open-source movie subtitles and we did some text pre-processing to filter out bad/inappropriate words before removed all of the diacritics for evaluation. This test corpus consists of 15,000 informal sentences.



    
\subsection{Experimental Setting}
The experiment of our model was implemented based on ALBERT \cite{lan2020albert} Transformer Encoders. We use LAMB \cite{you2020large} optimizer with learning rate of 1.76e-3 and the power of poly decay is 1.0. The model is trained for 500,000 steps with batch size equal to 512. Word and character vocabulary size are 60,000 and 400 respectively, which include the most common words and characters in the training data. In order to verify performance of our model, we trained three models, one is traditional word-level Bi-LSTM (Bidirectional long short-term memory) approach, one is Bi-LSTM with the combination of word and character, and an other is word-level Transformer Encoder to see the effectiveness of our hierarchical model. Bi-LSTM is a special kind of recurrent neural network, which are made of a set of forward LSTM and backward LSTM. LSTM first introduced in \cite{10.1162/neco.1997.9.8.1735} alleviates the vanishing gradient problem when captures long-term dependencies by having a set of memory blocks, each of the memory blocks has three different gates including input gate, forget gate, and output gate. Moreover, we also compare our proposed model with an external tool which is obtained from Viettel Group's NLP\footnote{\url{https://viettelgroup.ai/en/service/nlp}}.

Our experiments were evaluated using recall, precision and f1-score of the positive label (is-error label) for error detector. In order to measure correction performance, we computed accuracy score in two aspects:
\begin{itemize}
    \item Correction accuracy in the detected errors is formed by the ratio of exact correction and total correction.
        $$Accuracy_{in\ \%\ detected} = \frac{n_{exact\ correction}}{n_{exact\ correction} + n_{wrong\ correction}} $$
    \item Correction accuracy in total errors is counted by exact correction divided by the sum of total correction and wrong detection.
            $$Accuracy_{in\ total} = \frac{n_{exact\ correction}}{n_{exact\ correction} + n_{wrong\ correction} + n_{wrong\ detection}} $$
    
\end{itemize}

\subsection{Results}

The results of evaluating on the Wiki corpus are shown in Table \ref{table:2}. First, we observe that the hierarchical Transformer model themselves are quite useful (based on f-1 Score, correction accuracy on total and detected errors) to detect and suggest corrections for spelling errors. It achieves better scores in those metrics than both Bi-LSTM and external tool. Meanwhile, The Transformer model with multiple encoder layers (character and word) achieves higher scores on f1-score and correction accuracy on total errors than the Transformer model with encoders at word level but fails in correction accuracy on detected errors. To further clarify this, a higher detected recall score means the model predicts more errors than actual errors, which can lead to a low precision; therefore, we can observe that the Transformer model with char-word level encoders achieves a higher f-1 score. However, word level Transformer model is worse than char-word level Bi-LSTM model. The main reason is that the mistakes are not part of the vocabulary, and they will be split into sub-words and characters and use the shared embedding, meanwhile, combining character embedding and word embedding utilize more characteristics. Of those errors detected, both Transformer models suggest a high accuracy for spelling error corrections with 0.35\% higher for the model with encoders at word level. This difference is insignificant, since it is less than 1\%. Moreover, when weighting accuracy on total actual errors, Transformer model with multiple encoder layers at char-word levels achieves a significantly higher score than both Transformer model with encoders at word level and Bi-LSTM model with char-word layers (64.29\% compare to 33.28\% and 36.62\% respectively). The hierarchical Transformer model utilizes both character and word level encoders to detect Vietnamese spelling errors and make corrections outperformed related contemporary models (Transformer and Bi-LSTM) and the external tools available for the task.

In the subtitle corpus, the hierarchical Transformer with multiple encoders also proved to be efficacious in restore missing diacritics in short conversation text with 98.50\% in correction accuracy in the detected errors and 99.75\% in detector f1-score. Proving that our model is also practical for informal text. The results show in Table \ref{table:3}.
\def\arraystretch{1.5}%
\setlength{\tabcolsep}{0.72em} %

\begin{table}[htbp]
\centering
\caption{Evaluation  results in Vietnamese Spelling Corrections on Wiki Dataset}
\begin{tabular}{ p{1.7cm}|p{0.8cm}|P{1.3cm}|P{0.9cm}|P{0.7cm}|P{1.1cm}|P{2cm}  }
 \hline
 \multirow{2}{0pt}{\centering\textbf{Model}} & \multirow{2}{0pt}{\textbf{Level}} & \multicolumn{3}{c|}{\textbf{Detector}} & \multicolumn{2}{c}{\textbf{Corrector}}\\
\cline{3-7}
  &  & Precision & Recall & F1 & in total & in \% detected \\
 \hline

 & & & & & & \\
 Transformer Encoder (Ours) & Char-Word & \textbf{66.96} & \textbf{70.92} & \textbf{68.88} & \textbf{64.29} & \underline{96.01} \\
  & & & & & & \\
 Transformer Encoder & Word & 34.53 & 68.75 & 45.97 & 33.28 & \textbf{96.36} \\
  & & & & & & \\
 Bi-LSTM & Char-Word & \underline{39.12} &	\underline{68.76} &	\underline{49.87} &	\underline{36.62} &	93.59 \\
 & & & & & & \\
 Bi-LSTM & Word & 19.45 &	65.26 &	29.97 &	18.25 &	93.85 \\
  & & & & & & \\
  External Tool &  & 22.07 & 63.68 & 32.77 & 18.21 & 82.54 \\

 \hline
\end{tabular}
\small
\textit{Notes:}  The best scores are in bold; second best scores are underlined.
\label{table:2}
\end{table}
\def\arraystretch{1.5}%
\setlength{\tabcolsep}{0.72em} %

\begin{table}[htbp]
\centering
\caption{Evaluation results for the subtitle diacritics restoration}
\begin{tabular}{ p{1.7cm}|p{0.8cm}|P{1.3cm}|P{0.9cm}|P{0.7cm}|P{1.1cm}|P{2cm}  }
 \hline
 \multirow{2}{0pt}{\centering\textbf{Model}} & \multirow{2}{0pt}{\textbf{Level}} & \multicolumn{3}{c|}{\textbf{Detector}} & \multicolumn{2}{c}{\textbf{Corrector}}\\
\cline{3-7}
  &  & Precision & Recall & F1 & in total & in \% detected \\
 \hline

 & & & & & & \\
 Transformer Encoder (Ours) & Char-Word & \textbf{99.83} & \textbf{99.67} & \textbf{99.75} & \textbf{98.17} & \emph{98.50} \\
  & & & & & & \\
 Transformer Encoder & Word & \emph{99.70} & \emph{99.41} & \emph{99.56} & \emph{98.09} & \textbf{98.51} \\
  & & & & & & \\
 Bi-LSTM & Char-Word & 99.17 &	99.33 &	99.25 &	94.22 &	95.14 \\
 & & & & & & \\
 Bi-LSTM & Word & 98.92 & 97.51 & 98.21 & 95.25 & 96.72 \\
 \hline
\end{tabular}
\label{table:3}
\textit{Notes:}  The best scores are in bold; second best scores are underlined.
\end{table}

\section{Discussion and Future Work}
In this paper, we have proposed a hierarchical Transformer model for Vietnamese spelling error detection and correction. The model takes advantage of both character and context information in a hierarchical way to make predictions for each token of the input sequence. The model is trained over an extensive training set randomly generated from a corpus of high-quality news, Wiki, and subtitles. Through experiments, we have proved that our proposed model outperforms other architectures with significant margin. The evaluation is conducted under the Wiki spelling test set that we built. We have also employed the models into our products to support the users to write messages, news, and many other types of texts.

Although working well with user expectations, the model does not generally solve the spelling problem due to token-to-token mechanism. As we model the correction as classification problem and make only one prediction per token, the vocabulary of output for a single token is fixed. That means, for use-cases such as joined consecutive words (caused by white-space omission) or abbreviations, we could only set a limited number of phrases in the output vocabulary. In order to solve this use-case, we tested with machine translation model as in \cite{Vaswani2017AttentionIA} but the inference speed is dramatically slower and not worth the response time.

For future work, we plan to expand the Wikipedia spelling test further and build a similar set for typical Vietnamese news and verbal messages. We also want to design more robust and powerful architectures that allow us to offer a more satisfying experience to users.



\renewcommand\bibname{References}
\bibliography{main.bbl}

\begin{thebibliography}{10}
\providecommand{\url}[1]{\texttt{#1}}
\providecommand{\urlprefix}{URL }
\providecommand{\doi}[1]{https://doi.org/#1}

\bibitem{Devlin2019BERTPO}
Devlin, J., Chang, M.W., Lee, K., Toutanova, K.: Bert: Pre-training of deep
  bidirectional transformers for language understanding. In: NAACL-HLT (2019)

\bibitem{10.5555/177910.177914}
Gage, P.: A new algorithm for data compression. C Users J.  \textbf{12}(2),
  23–38 (Feb 1994)

\bibitem{10.1162/neco.1997.9.8.1735}
Hochreiter, S., Schmidhuber, J.: Long short-term memory. Neural Comput.
  \textbf{9}(8),  1735–1780 (Nov 1997). \doi{10.1162/neco.1997.9.8.1735},
  \url{https://doi.org/10.1162/neco.1997.9.8.1735}

\bibitem{lan2020albert}
Lan, Z., Chen, M., Goodman, S., Gimpel, K., Sharma, P., Soricut, R.: Albert: A
  lite bert for self-supervised learning of language representations (2020)

\bibitem{10.1007/978-981-15-6168-9_40}
Nguyen, H.T., Dang, T.B., Nguyen, L.M.: Deep learning approach for vietnamese
  consonant misspell correction. In: Nguyen, L.M., Phan, X.H., Hasida, K.,
  Tojo, S. (eds.) Computational Linguistics. pp. 497--504. Springer Singapore,
  Singapore (2020)

\bibitem{4586339}
{Nguyen}, P.H., {Ngo}, T.D., {Phan}, D.A., {Dinh}, T.P.T., {Huynh}, T.Q.:
  Vietnamese spelling detection and correction using bi-gram, minimum edit
  distance, soundex algorithms with some additional heuristics. In: 2008 IEEE
  International Conference on Research, Innovation and Vision for the Future in
  Computing and Communication Technologies. pp. 96--102 (2008)

\bibitem{Niranjan2020HierarchicalAT}
Niranjan, A., Shaik, M.A.B., Verma, K.: Hierarchical attention transformer
  architecture for syntactic spell correction. ArXiv  \textbf{abs/2005.04876}
  (2020)

\bibitem{korean}
Park, C., Kim, K., Yang, Y., Kang, M., Lim, H.: Neural spelling correction:
  translating incorrect sentences to correct sentences for multimedia.
  Multimedia Tools and Applications  (2020). \doi{10.1007/s11042-020-09148-2}

\bibitem{DBLP:journals/corr/abs-1709-07104}
Pham, T., Pham, X., Le{-}Hong, P.: On the use of machine translation-based
  approaches for vietnamese diacritic restoration. CoRR
  \textbf{abs/1709.07104} (2017), \url{http://arxiv.org/abs/1709.07104}

\bibitem{Vaswani2017AttentionIA}
Vaswani, A., Shazeer, N., Parmar, N., Uszkoreit, J., Jones, L., Gomez, A.N.,
  Kaiser, L., Polosukhin, I.: Attention is all you need. ArXiv
  \textbf{abs/1706.03762} (2017)

\bibitem{wu2016googles}
Wu, Y., Schuster, M., Chen, Z., Le, Q.V., Norouzi, M., Macherey, W., Krikun,
  M., Cao, Y., Gao, Q., Macherey, K., Klingner, J., Shah, A., Johnson, M., Liu,
  X., Łukasz Kaiser, Gouws, S., Kato, Y., Kudo, T., Kazawa, H., Stevens, K.,
  Kurian, G., Patil, N., Wang, W., Young, C., Smith, J., Riesa, J., Rudnick,
  A., Vinyals, O., Corrado, G., Hughes, M., Dean, J.: Google's neural machine
  translation system: Bridging the gap between human and machine translation
  (2016)

\bibitem{you2020large}
You, Y., Li, J., Reddi, S., Hseu, J., Kumar, S., Bhojanapalli, S., Song, X.,
  Demmel, J., Keutzer, K., Hsieh, C.J.: Large batch optimization for deep
  learning: Training bert in 76 minutes (2020)

\bibitem{Zhang2020SpellingEC}
Zhang, S., Huang, H., Liu, J., Li, H.: Spelling error correction with
  soft-masked bert. In: ACL (2020)

\end{thebibliography}

\end{document}